\documentclass[10pt,twocolumn,letterpaper]{article}

\usepackage{iccv}              
\usepackage{graphicx}	
\usepackage{amsmath}	
\usepackage{amssymb}	
\usepackage{booktabs}
\usepackage{times}
\usepackage{microtype}
\usepackage{epsfig}
\usepackage{caption}
\usepackage{float}
\usepackage{placeins}
\usepackage{color, colortbl}
\usepackage{stfloats}
\usepackage{enumitem}
\usepackage{tabularx}
\usepackage{xstring}
\usepackage{multirow}
\usepackage{xspace}
\usepackage{url}
\usepackage{subcaption}
\usepackage{xcolor}
\usepackage[hang,flushmargin]{footmisc}

\newcommand{\ziquan}[1]{\textcolor{black}{#1}}

\newcommand{\R}[1]{{%
    \textbf{%
        \ifstrequal{#1}{1}{\textcolor{red}{R#1}}{%
        \ifstrequal{#1}{2}{\textcolor{blue}{R#1}}{%
        \ifstrequal{#1}{3}{\textcolor{magenta}{R#1}}{%
        \ifstrequal{#1}{4}{\textcolor{teal}{R#1}}{%
                           \textcolor{cyan}{R#1}%
        }}}}%
    }%
}}
\definecolor{iccvblue}{rgb}{0.21,0.49,0.74}
\usepackage[pagebackref,breaklinks,colorlinks,allcolors=iccvblue]{hyperref}

\def\paperID{6210} 
\def\confName{ICCV}
\def\confYear{2025}

\title{ConformalSAM: Unlocking the Potential of Foundational Segmentation Models in Semi-Supervised Semantic Segmentation with Conformal Prediction}

\author{
  Danhui Chen\textsuperscript{1}\thanks{Equal contribution. \textsuperscript{$\dagger$}Corresponding author.} \quad
  Ziquan Liu\textsuperscript{2}\footnotemark[1] \quad
  Chuxi Yang\textsuperscript{1}\footnotemark[1] \quad
  Dan Wang \textsuperscript{1} \\ \quad
  Yan Yan\textsuperscript{3} \quad
  Yi Xu\textsuperscript{1$\dagger$} \quad
  Xiangyang Ji \textsuperscript{4} \\
  \textsuperscript{1}Dalian University of Technology \quad
  \textsuperscript{2}Queen Mary University of London \quad \\
  \textsuperscript{3}Washington State University \quad
  \textsuperscript{4}Tsinghua University\\
  {\tt\small darrenchen0104@gmail.com, ziquan.liu@qmul.ac.uk, wangdan@mail.dlut.edu.cn,}\\ {\tt\small \{yangcx, yxu\}@dlut.edu.cn, yan.yan1@wsu.edu, xyji@tsinghua.edu.cn}
}

\begin{document}
\maketitle
\begin{abstract}
Pixel-level vision tasks, such as semantic segmentation, require extensive and high-quality annotated data, which is costly to obtain. Semi-supervised semantic segmentation (SSSS) has emerged as a solution to alleviate the labeling burden by leveraging both labeled and unlabeled data through self-training techniques. Meanwhile, the advent of foundational segmentation models pre-trained on massive data, has shown the potential to generalize across domains effectively. This work explores whether a foundational segmentation model can address label scarcity in the pixel-level vision task as an annotator for unlabeled images. Specifically, we investigate the efficacy of using SEEM, a Segment Anything Model (SAM) variant fine-tuned for textual input, to generate predictive masks for unlabeled data. To address the shortcomings of using SEEM-generated masks as supervision, we propose ConformalSAM, a novel SSSS framework which first calibrates the foundation model using the target domain's labeled data and then filters out unreliable pixel labels of unlabeled data so that only high-confidence labels are used as supervision. By leveraging  conformal prediction (CP) to adapt foundation models to target data through uncertainty calibration, ConformalSAM exploits the strong capability of the foundational segmentation model reliably which benefits the early-stage learning, while a subsequent self-reliance training strategy mitigates overfitting to SEEM-generated masks in the later training stage. Our experiment demonstrates that, on three standard benchmarks of SSSS, ConformalSAM achieves superior performance compared to recent SSSS methods and helps boost the performance of those methods as a plug-in.
\end{abstract}    
\section{Introduction}
\label{sec:intro}

\begin{figure}[t]
    \centering
    \includegraphics[width=0.4\textwidth]{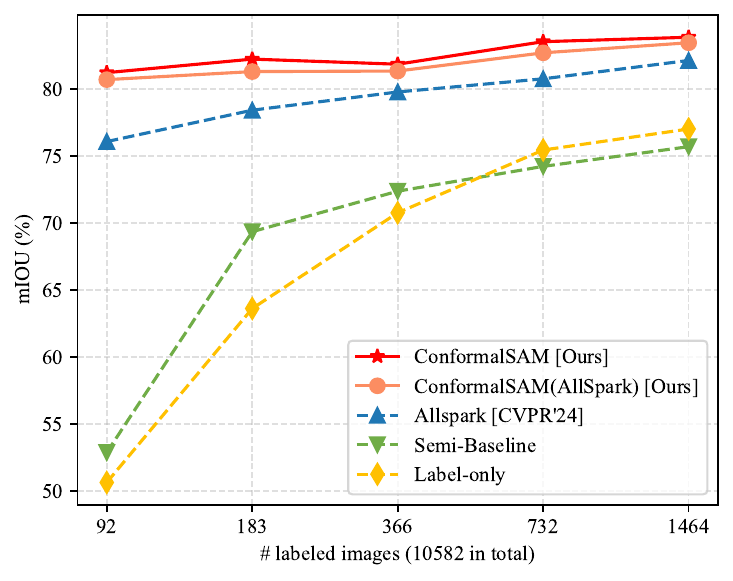}
    \vspace{-0.4cm}
    \caption{Comparison of our method (ConformalSAM) with state-of-the-art methods on the PASCAL VOC dataset \cite{everingham2015pascal}. ConformalSAM achieves the best performance across all splits. ConformalSAM (AllSpark), a variant using the AllSpark \cite{wang2024allspark} framework, also shows significant improvements.}
    \label{fig:fig-1.png}
    \vspace{-0.6cm}
\end{figure}

\ziquan{One of the long-standing challenges in pixel-level vision tasks, such as semantic segmentation, is balancing the need for high-quality mask annotations with the high costs associated with labeling. To address this problem, semi-supervised semantic segmentation (SSSS) \cite{chen2021semi,tarvainen2017mean,zoupseudoseg,yang2023revisiting,wang2024allspark} has been proposed to train high-quality segmentation models with limited labels. Existing SSSS methods mainly focus on designing better self-training strategies, such as Mean-teacher \cite{tarvainen2017mean} and Cross Pseudo
Supervision \cite{chen2021semi}, to maximize the use of both labeled and unlabeled samples. Meanwhile, foundation models pre-trained on massive amounts of data such as CLIP \cite{radford2021learning} and SAM (Segment Anything Model) \cite{kirillov2023segment} have transformed computer vision through their highly generalizable representations. This naturally raises the question: can the strong capability of foundation models address the label scarcity issue in pixel-level vision understanding and improve the efficacy of SSSS? }

\begin{figure*}
    \centering
    \vspace{-0.6cm}
    \includegraphics[width=0.9\textwidth]{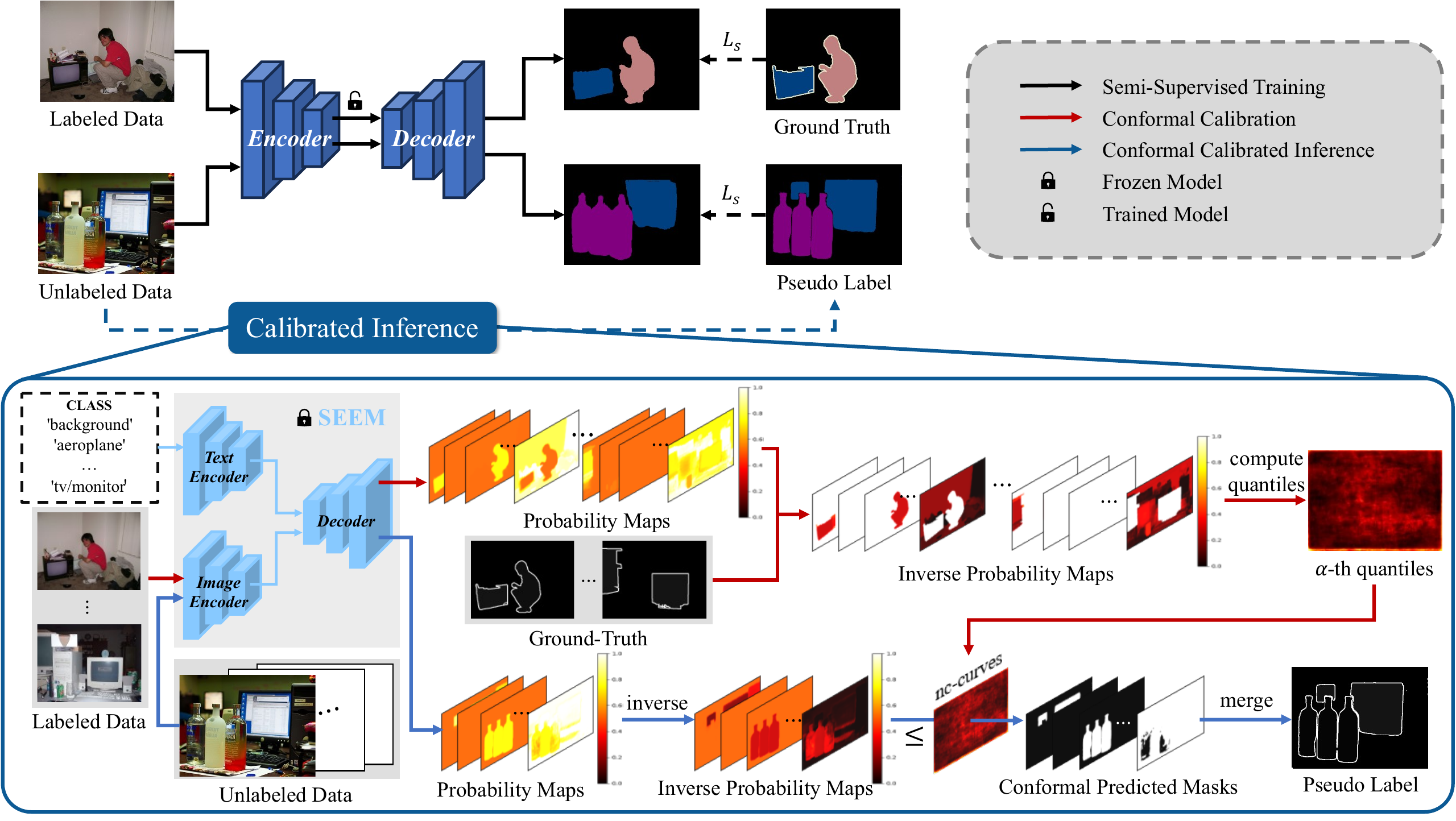}
    \vspace{-0.2cm}
    \caption{Overview of ConformalSAM. We adopt the simple self-training framework as our basic framework (top), where labeled and unlabeled data are supervised by ground-truth and pseudo labels respectively. The key difference lies in the generation of pseudo labels: instead of using the model to produce pseudo labels as in traditional self-training, we leverage an uncertainty-calibrated SEEM model to generate segmentation masks (bottom). Conformal Calibration (red arrows in the lower section) is the process of calibrating the SEEM with labeled data to obtain the non-conformity curve. Conformal Calibrated Inference (blue arrows in the lower section) then uses this non-conformity curve to generate pseudo labels for unlabeled data.}\label{fig:overview}
    \vspace{-0.5cm}
\end{figure*}

\ziquan{To answer this question, we investigate the feasibility of using predictive masks for unlabeled samples generated by SEEM \cite{zou2023segment}, a variant of SAM fine-tuned to handle textual input, as their supervision in SSSS. Different from existing works that either fine-tune a foundation segmentation model \cite{miao2024cross} or incorporate their predictive masks in classical SSSS methods such as consistency \cite{zhang2023semisam}, we adopt the foundation model guided by Occam's razor: we train our segmentation model directly on both human-labeled samples and SEEM-labeled samples. Not surprisingly, directly using the predictive masks from SEEM does not work well in SSSS as the domain gap between SEEM's pre-training data and the target data leads to low-quality pixel labels. To unleash the power of the foundational segmentation model in a target domain when there are limited labels, we propose \textbf{\emph{ConformalSAM}}, an SSSS framework based on conformal prediction (CP) \cite{shafer2008tutorial,angelopoulos2020uncertainty,angelopoulos2021gentle,brunekreef2024kandinsky}. ConformalSAM first calibrates a foundational segmentation model using the target domain's labeled data and then uses the calibrated predictive labels as the supervision for unlabeled data to train a performant model in the target domain. We select conformal prediction (CP) due to its black-box nature, allowing it to estimate uncertainty in a foundation model with only limited labeled data, thereby enabling the model to focus on the generation of relevant knowledge for the target domain. The uncertainty information is employed to produce high-confidence labels that are valuable for early-stage learning in the target domain, especially for identifying non-background classes when the background pixels are dominant. In contrast, later-stage training has the risk of overfitting the model to SEEM's masks, so we use a self-reliance transition strategy that discards SEEM's masks and only uses self-training to refine the target model. }

\ziquan{Our ConformalSAM achieves better performance than recent SSSS methods as it strikes a balance between exploiting the strong generalization ability of foundation models and leveraging the domain data. Our experiment shows that CP substantially improves the quality of SEEM masks and releases the potential of the foundation model even without self-reliance training, revealing the effectiveness of CP in filtering out unreliable predictions. When combined with the later-stage self-reliance training, the performance of ConformalSAM is further improved. Fig.~\ref{fig:fig-1.png} shows that ConformalSAM performs better than a recent strong baseline AllSpark \cite{wang2024allspark}. Moreover, when combined with AllSpark, ConformalSAM improves upon its performance, showing the potential of ConformalSAM in existing SSSS methods. Our key contributions are as follows:}

\begin{itemize}
\item[$\bullet$] We investigate the effectiveness of the foundational segmentation model, SEEM, as a labeling tool for unlabeled samples in a downstream SSSS task. Our findings reveal the limitations and weak performance of this straightforward approach to using foundational models.
\item[$\bullet$] Our ConformalSAM is among the first to use CP to overcome the difficulty of using a foundational segmentation model as the labeler in SSSS. With the help of CP, ConformalSAM probes the uncertainty of the foundation model in a target domain with labeled training data so that high-confidence masks are selected as supervision labels. In addition, for a target task with the background class, we design a class-conditional filtering based on CP to avoid the overwhelming background pixels. 

\item[$\bullet$] The effectiveness of ConformalSAM is demonstrated through extensive experiments, showing improved performance over recent SSSS baselines on three standard datasets: PASCAL VOC, PASCAL VOC augmented and ADE20K. Furthermore, as a versatile plug-in framework, ConformalSAM boosts performance when integrated with existing SSSS methods, underscoring its adaptability and effectiveness. Specifically, ConformalSAM outperforms AllSpark \cite{wang2024allspark} by an average of 3.37\% across all SSSS experimental settings; when combined with ConformalSAM, Allspark’s average performance improves by 2.07\%.
\end{itemize}
\section{Related Work}
\label{sec:related}
\vspace{-0.2cm}
\noindent\textbf{Foundational Segmentation Models in Downstream Tasks.} SAM \cite{kirillov2023segment} is a visual foundation model for segmentation that has high-quality segmentation results and outstanding performance in zero-shot segmentation scenarios. SAM has been leveraged in different downstream tasks to tackle complex visual challenges effectively. S2C \cite{kweon2024sam} transfers SAM's knowledge to the classifier during training, improving class activation maps (CAMs). WS-SAM \cite{he2024weakly} uses sparse annotations as SAM prompts to address weak supervision challenges. RSPrompter \cite{chen2024rsprompter} learns to generate prompts, enhancing SAM's performance in remote sensing image segmentation. CPC-SAM \cite{miao2024cross} introduces a cross-prompting strategy in a dual-branch framework for automatic prompt generation. SemiSAM \cite{zhang2023semisam} leverages domain knowledge for localization, using SAM-generated pseudo-labels as additional supervision in a semi-supervised framework.
Our work is different from the existing works in three aspects: 1) existing approaches mostly focus on generating better prompts to drive SAM for segmentation, whereas we directly use SEEM to generate masks for unlabeled images in SSSS, 2) we use CP to efficiently and effectively ground the knowledge of the foundation model to the downstream tasks, 3) we propose a two-stage training strategy to balance the use of the foundation model and the downstream data.

\noindent\textbf{Semi-Supervised Semantic Segmentation.}
Earlier semi-supervised semantic segmentation methods \cite{chen2021semi,tarvainen2017mean,zoupseudoseg} have typically been investigated in terms of how to better utilize unlabeled data, and most of these approaches have used CNN-based backbones. One representative work is UniMatch \cite{yang2023revisiting}, which proposes a Unified Dual-Stream Perturbations approach that extends the perturbation space and sufficiently exploits image-level perturbations by introducing auxiliary feature perturbation stream and dual-stream perturbation techniques. AllSpark \cite{wang2024allspark} extends SSSS further based on Transformer, which solves the problem of labeled data dominating the training process, by introducing unlabeled features among labeled features through a channel-wise cross-attention mechanism.

\noindent\textbf{Conformal Prediction.} We briefly introduce the concept of conformal prediction \cite{papadopoulos2002inductive,vovk2005algorithmic,lei2014distribution,shi2025direct,shi2024conformal,liu2024pitfalls}. Let us consider a classification task first. Given a training set $D_{train}$, and the predictive model $\hat{f}$ that have been trained on this training set, the model's output for an input image $x$ is $f(x) \in [0,1]^k$, where $k$ is the number of classes. As the first step of conformal prediction, we use a calibration set $D_{cal}$ to compute a threshold of non-conformity scores to determine the prediction set size during inference. The calibration set consists of data disjoint from the training set and is written as: $D_{cal} = \{(x_i, y_i)\}_{i=1}^n$. First, we compute the non-conformity score for each sample in the calibration set. One possible non-conformity score is: $s_i = 1 - \hat{f}_{y_i}(x_i)$, where $\hat{f}_{j}(\cdot)$ represents the predicted probability of the model for the ground-truth class $j$ for sample $x_i$. Next, we sort the non-conformity scores $[s_1, s_2, \dots, s_n]$, and compute the $\lceil(n+1)(1-\alpha)/n\rceil$ quantile, denoted by $\hat{q}$, where $\lceil\cdot\rceil$ means round up, $\alpha$ is the predefined tolerance rate. Based on this quartile $\hat{q}$, we define the prediction set for a novel test image $x_{test}$ as: 
$C(x_{test}) = \left\{ j : 1 - \hat{f}_{j}(x_{test}) \leq \hat{q} \right\}$. The conformal prediction ensures that the prediction set satisfies the following coverage:
\setlength{\belowdisplayskip}{1.0pt} \setlength{\belowdisplayshortskip}{1.0pt}
\setlength{\abovedisplayskip}{1.0pt} \setlength{\abovedisplayshortskip}{1.0pt}
\begin{align}
1 - \alpha \leq P(y_{\text{test}} \in C(x_{\text{test}})) \leq 1 - \alpha + \frac{1}{n+1}.
\end{align}
This equation demonstrates that conformal prediction provides prediction sets with a theoretically guaranteed coverage rate. As a result, conformal prediction is particularly useful in applications requiring reliable prediction results. 

In semantic segmentation, each pixel in an image is individually assigned to a specific class, treating each pixel's classification as part of a unified segmentation process rather than independent tasks. Conformal prediction has been applied to segmentation tasks in \cite{brunekreef2024kandinsky}. To balance the trade-off between pixel-level calibration precision and data availability, an intermediary method named Kandinsky-calibration \cite{brunekreef2024kandinsky} has been proposed, which leverages prior knowledge of spatial correlations to aggregate globally comparable pixels for calibration. Our work is different from \cite{brunekreef2024kandinsky} as we aim to tackle the challenge of using pseudo-labels generated by a foundational segmentation model in the SSSS task and proposes a novel class-conditioned CP for balancing background and object classes. Furthermore, our empirical study demonstrates the superiority of using pixel-wise CP in our setting compared with the Kandinsky CP proposed in \cite{brunekreef2024kandinsky}.
\vspace{-0.3cm}
\section{Method}
\label{sec:method}
\vspace{-0.1cm}
This section begins by presenting the preliminaries through a formal definition of the semi-supervised semantic segmentation problem (Section \ref{Preliminaries}). Next, in Section \ref{SEEM}, we present our initial approach to applying SEEM to semi-supervised semantic segmentation (SSSS) and discuss the challenges encountered. Finally, we detail our proposed two-stage method, including the use of CP for calibrating the foundation model in Stage I (Section \ref{CP})and the Self-Reliance Training strategy introduced in Stage II(Section \ref{decay}).
\subsection{Preliminaries}\label{Preliminaries}
In the SSSS task, the dataset $D$ is composed of a small set of labeled data $D_l = \{(x_i, y_i)\}_{i=1}^L$ and a large set of unlabeled data $D_u = \{(x_i)\}_{i=1}^U$, where $ x_i \in \mathbb{R}^{3 \times H \times W}$ represents the RGB image input, and \( y_i \in \mathbb{R}^{K \times H \times W} \) is the corresponding ground truth label. Here, \( K \) represents the number of segmentation classes, and \( L \) and \( U \) denote the number of labeled and unlabeled data points, respectively, often with \( L \ll U \). Our objective is to leverage the small amount of labeled data and the large amount of unlabeled data to train a performant semantic segmentation model. Based on pseudo-labeling methods, current research typically involves first training a model on labeled data and then using the trained model to generate pseudo-labels \( y_j \) for the unlabeled data. These pseudo-labels are then used as supervision labels for the next round of training. The objective function for this method is:
\begin{align}
\mathcal{L} = \frac{1}{L} \sum_{i=0}^{L} \mathcal{L}_{CE}(f_\theta(x_i), y_i) + \lambda \times \frac{1}{U} \sum_{j=0}^{U} \mathcal{L}_{CE}(f_\theta(x_j), \hat{y}_j)
\label{eqn:loss}
\end{align}
where \(L_{CE}\) is the cross-entropy loss, \( \theta \) denotes the model parameters to be optimized, and \( \lambda \) controls the contribution of the unsupervised loss.

\subsection{Effect of SEEM-Generated Masks in SSSS}\label{SEEM}
We first explore the effectiveness of directly using pseudo-masks generated by SEEM. The reason why we use SEEM instead of SAM is that SEEM allows textual prompts as input. For each dataset, we input $K$ class names and an unlabeled image to SEEM to obtain $K$ masks. For each pixel, we set its label as the class with the maximum score. Given the SEEM-generated masks, we train a segmentation model with ground-truth masks for labeled images and SEEM-generated masks for unlabeled images. We compare this model with a model trained with ground-truth labels, i.e., only the first term in Eqn.~\ref{eqn:loss}, in Tab.~\ref{tab:seem_vanilla}. It turns out that using SEEM-generated labels as the supervision directly often leads to worse performance in both datasets, indicating that a foundation model is not always helpful to downstream tasks. The SEEM-generated and ground-truth masks are shown in Fig.~\ref{fig:visualization}. revealing that the SEEM-generated masks are low-quality. 

\begin{table}[t]
\renewcommand{\arraystretch}{1.2}
\centering (a) PASCAL VOC\\
\resizebox{1.0\linewidth}{!}{ 
\small
\begin{tabular}{c|ccccc}
\hline
Method        & 1/16(92) & 1/8(183) & 1/4(366) & 1/2(732)  \\ \hline
Label-only    & \textbf{50.65}    & \textbf{63.62}    & \textbf{70.76}    & \textbf{75.44}     \\
Label+Unlabel(SEEM)      & 42.00    & 41.90    & 42.37    & 44.99     \\
\hline
\end{tabular}
}
\centering (b) ADE20K\\
\resizebox{1.0\linewidth}{!}{ 
\begin{tabular}{c|ccccc}
\hline
Method        & 1/64(316) & 1/32(632) & 1/16(1263) & 1/8(2526)  \\ \hline
Label-only       & 18.83    & \textbf{24.77}    & \textbf{28.76}  & \textbf{32.67}   \\
Label+Unlabel(SEEM)       & \textbf{20.53}    & 21.01    & 21.45 & 22.03    \\
\hline
\end{tabular}
}
\vspace{-0.2cm}
\caption{Using pseudo-labels generated by SEEM for unlabeled data in PASCAL VOC \cite{everingham2015pascal} and ADE20K \cite{zhou2017scene}. The first row is the ratio of labeled to unlabeled data and the rest two are mIOU. }
\vspace{-0.5cm}
\label{tab:seem_vanilla}
\end{table}

\subsection{ConformalSAM}
A significant challenge with SEEM-generated masks is their inconsistent quality, stemming from the downstream task's detachment from the foundation model. This detachment is reflected in the low predictive uncertainty of SEEM at some pixel locations, so directly using the predictive class as the label for those pixels can lead to misleading supervision and suboptimal results. To exploit the foundation model's ability in a specific domain, we propose ConformalSAM that uses an uncertainty-aware method, conformal prediction (CP), with samples in the labeled set to filter out unreliable predictions. This calibration process is consistent with the prediction-power prediction \cite{angelopoulos2023prediction} where ground-truth labels are used to debias a machine learning model during inference. Here we use the labeled set as the calibration set and obtain debiased pseudo-labels from SEEM for supervised training in SSSS. The second step of ConformalSAM is to switch from SEEM-assisted training to self-reliance training, when SEEM-generated masks are dropped and only pseudo-labels from the downstream model are used. The remainder of this section describes the two stages in detail.

\subsubsection{CP-Calibrated Foundation Model for Supervised Training (Stage \uppercase\expandafter{\romannumeral1})}\label{CP}

\begin{figure}[ht]
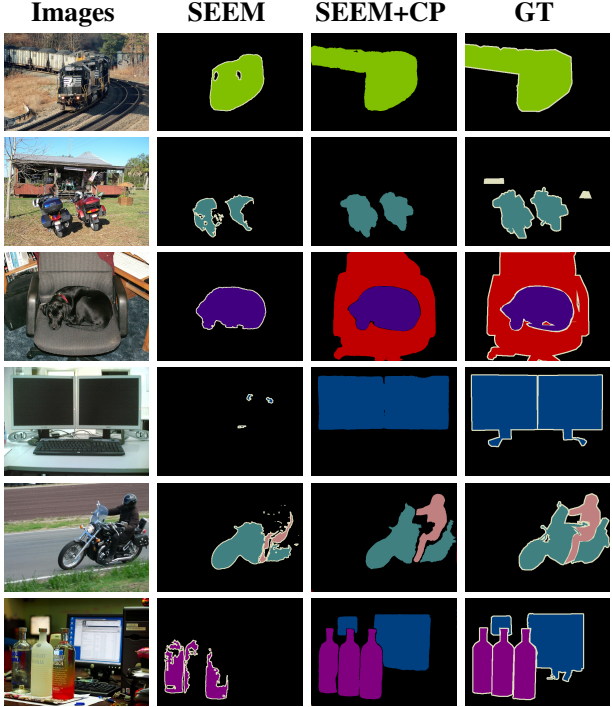

    \centering
    \vspace{-0.3cm}
    \renewcommand{\arraystretch}{1.5} 
    \scalebox{0.5}{ 
        \begin{tabular}{>{\centering\arraybackslash}m{0.21\textwidth} >{\centering\arraybackslash}m{0.21\textwidth} >{\centering\arraybackslash}m{0.21\textwidth} >{\centering\arraybackslash}m{0.21\textwidth}} 
            \huge \textbf{Images} &  \huge \textbf{SEEM} &  \huge \textbf{SEEM+CP} &  \huge \textbf{GT} \\[5pt]
            \includegraphics[width=0.22\textwidth]{train.pdf} &
            \includegraphics[width=0.22\textwidth]{train_SEEM.pdf} &
            \includegraphics[width=0.22\textwidth]{train_cp.pdf} &
            \includegraphics[width=0.22\textwidth]{train_gt.pdf} \\

            \includegraphics[width=0.22\textwidth]{mt.pdf} &
            \includegraphics[width=0.22\textwidth]{mt_SEEM.pdf} &
            \includegraphics[width=0.22\textwidth]{mt_cp.pdf} &
            \includegraphics[width=0.22\textwidth]{mt_gt.pdf} \\

            \includegraphics[width=0.22\textwidth]{dog.pdf} &
            \includegraphics[width=0.22\textwidth]{dog_SEEM.pdf} &
            \includegraphics[width=0.22\textwidth]{dog_cp.pdf} &
            \includegraphics[width=0.22\textwidth]{dog_gt.pdf} \\

            \includegraphics[width=0.22\textwidth]{tv.pdf} &
            \includegraphics[width=0.22\textwidth]{tv_SEEM.pdf} &
            \includegraphics[width=0.22\textwidth]{tv_cp.pdf} &
            \includegraphics[width=0.22\textwidth]{tv_gt.pdf} \\

            \includegraphics[width=0.22\textwidth]{mtr.pdf} &
            \includegraphics[width=0.22\textwidth]{mtr_SEEM.pdf} &
            \includegraphics[width=0.22\textwidth]{mtr_cp.pdf} &
            \includegraphics[width=0.22\textwidth]{mtr_gt.pdf} \\

            \includegraphics[width=0.22\textwidth]{bottle.pdf} &
            \includegraphics[width=0.22\textwidth]{bottle_SEEM.pdf} &
            \includegraphics[width=0.22\textwidth]{bottle_cp.pdf} &
            \includegraphics[width=0.22\textwidth]{bottle_gt.pdf} \\
        \end{tabular}
    }
\vspace{-0.3cm}
    \caption{Segmentation masks from the PASCAL VOC dataset. Different columns display the original images, segmentation results from the SEEM model, SEEM segmentation results with conformal calibration (SEEM+CP), and ground truth labels. SEEM tends to mistakenly label object pixels as background and the class-conditional CP fixes this issue.}
    \label{fig:visualization}
    \vspace{-0.6cm}
\end{figure}

The first step in this stage is to generate a threshold for prediction using the calibration set, or the labeled images in SSSS. We use the given labeled data $D_l$ as the calibration set and perform CP on unlabeled data to obtain the calibrated prediction. 
As shown in Fig.~\ref{fig:overview}, for any image $(x_i, y_i) \in D_l, i = 1, \dots, L$ in the calibration set, we use the pre-trained SEEM $M_{\text{SEEM}}$ to generate the \ziquan{predictive probability} map of the image $P_i \in \mathbb{R}^{K \times H \times W}$. where the range of each element is $[0,1]$. Then we obtain the inverse prediction map $\hat{P}_i \in \mathbb{R}^{K \times H \times W}$:
\begin{equation}
\hat{P}_i^j(a, b) = 
\begin{cases} 
1 - P_i^j(a, b), & \text{if } y_i^j(a, b) = 1 \\ 
\text{NaN}, & \text{otherwise}
\end{cases}\label{equ_3}
\end{equation} 
\noindent{Here, $P_i^j(a, b)$ represents the value at the pixel location $(a, b)$ in the $j$-th channel of the predicted segmentation map. According to Eqn. \ref{equ_3}, we can obtain the non-conformity scores for different categories at each pixel in each image. Based on the pixel-level classification requirements, we combine the non-conformity scores of all categories at the same pixel location from every image in $D_c$ to get $\hat{S}(a, b) \in \mathbb{R}^{L \cdot K}$. Finally, we calculate the $(1-\alpha)$-quantile of the non-conformity score for each pixel location, obtaining $\hat{q}_\alpha \in \mathbb{R}^{H \times W}$.}

To get high-quality masks from SEEM, we use CP to filter out unreliable predictions from $M_{\text{SEEM}}$. Specifically, for an image $x_i$ in $D_u$, using the class name as a prompt for $M_{\text{SEEM}}$ we obtain the predicted probability map $P_i$ and its inverse $\hat{P}_i$. For each pixel of an unlabeled image $x_i$, we have the prediction set $\mathcal{C}_i(a,b)$ where $j$ belongs to $\mathcal{C}_i(a,b)$ if and only if $\hat{P}_{i}^{j}(a, b) \leq \hat{q}_{\alpha}(a, b)$. 
To obtain a calibrated segmentation mask $M_i$ for $x_i$, we use the following function
\begin{equation}
\tiny
M_i(a, b)=
\begin{cases} 
\arg\min_j \mathcal{C}_i(a,b)[j],& |\mathcal{C}_i(a,b)|> 0\And 0 \notin \mathcal{C}_i(a,b)\\
\arg\min_{j \neq 0} \mathcal{C}_i(a,b)[j], &|\mathcal{C}_i(a,b)|> 0\And 0 \in \mathcal{C}_i(a,b)\\
\text{NaN},& |\mathcal{C}_i(a,b)|=0
\end{cases}
\label{eqn:equ_5}
\end{equation}
when 0 is a background class. In a dataset with a background class such as PASCAL VOC \cite{everingham2015pascal}, the background class dominates over other classes, so we choose to give an object label to a pixel even though the background class has higher confidence. The class-conditional filtering favors minority classes over the majority class, which is motivated by \cite{shi2024conformal} revealing the class-wise coverage in CP is non-trivial for minority classes. When there is no background class such as in ADE20K \cite{zhou2017scene}, we do not consider the second line in Eqn.~\ref{eqn:equ_5}. In the experiment section, we analyze the effect of these components and the effect of class-conditional label selection is shown in Fig.~\ref{fig:visualization}. 

\subsubsection{Self-Reliance Training (Stage \uppercase\expandafter{\romannumeral2})}\label{decay}
Although the previous section \ref{CP} demonstrates that CP effectively improves the quality of pseudo-labels generated by SEEM, an unavoidable discrepancy remains between the refined pseudo-labels and the ground truth. In the stage I, this discrepancy is acceptable, as we aim to extract valuable information from the highly generalizable representation of the foundation model. However, leaving these differences unaddressed would significantly impact the model's final performance. To address this issue, we drop SEEM-generated masks in Stage II and employ a self-reliance training mechanism to direct the model's focus toward the downstream task. Meanwhile, self-reliance training may also adversely impact the model by reinforcing incorrect pseudo-labels, especially given the substantial error accumulation from pre-training and the scarcity of labeled data, amplifying this effect. To address this issue, several existing methods, such as FixMatch \cite{sohn2020fixmatch}, AEL \cite{hu2021semi}, and DAW \cite{sun2024daw}, employ dynamically adjusted thresholds during pseudo-label training to filter out potentially incorrect pseudo-labels, thereby mitigating the risk of model overfitting.

We propose a simple dynamic weighting strategy that adjusts the balance between ground-truth label supervised and pseudo-label unsupervised loss components to mitigate overfitting in the late stage.
\vspace{-0.1cm}
\begin{equation}
\mathcal{L} = (1-\lambda(t)) \times \mathcal{L}_{s} + \lambda(t) \times \mathcal{L}_{u}
\label{equ_6}
\end{equation} 
Specifically, the total loss function for training can be defined in Eqn. \ref{equ_6}, $\mathcal{L}_{s}$ and $\mathcal{L}_{u}$ denote the supervised and unsupervised loss, respectively. $\lambda(t)$ is a dynamically adjusted weight that varies with the training epoch. Although the relationship between supervised and unsupervised loss may be more complex, a simple exponential decay strategy leads to significant performance gains in our experiment. By gradually decreasing $\lambda$ during training, we ensure its influence on the model diminishes as the optimization progresses. This allows the model to rely increasingly on reliable supervised signals in the later stages, leading to more robust training outcomes. Our strategy not only mitigates the negative effects of pseudo-label but also significantly suppresses overfitting, thereby enhancing the model's generalization performance.

ConformalSAM's self-training framework is flexible and can be replaced with other methods, such as AllSpark, creating \textit{\textbf{ConformalSAM(AllSpark)}}. Specifically, in \textbf{Stage I}, we supervise the unlabeled data using pseudo-labels generated through calibrated inference, rather than pseudo-labels generated by the target model as in AllSpark. In \textbf{Stage II}, we discard the SEEM-generated labels in Section~\ref{CP} and incorporate the AllSpark framework, applying the dynamic loss weight adjustment strategy discussed earlier to further improve the training process. See the overview of ConformalSAM (AllSpark) in the Appendix.
\begin{table*}[]
\renewcommand{\arraystretch}{1.2}
\vspace{-0.5cm}
\centering
\begin{tabular}{c|c|ccccc}
\hline
\rowcolor[HTML]{DAE8FC} 
Method                                     & Encoder & 1/16(92)                     & 1/8(183)                     & 1/4(366)                     & 1/2(732)                     & Full(1464)                   \\ \hline

Label-only                     & SegFormer-B5     & 50.65                        & 63.62                        & 70.76                        & 75.44                        & 77.01                        \\ \hline
CutMix-Seg \cite{French2019SemisupervisedSS}\ \textcolor[HTML]{9B9B9B}{\scriptsize[{BMVC'20}]} & ResNet-101     & 55.58               & 63.20                 & 68.36               & 69.84          & 76.54       \\
CPS \cite{chen2021semi}\ \textcolor[HTML]{9B9B9B}{\scriptsize[{CVPR'21}]} &ResNet-101     & 64.07              & 67.42                 & 71.71                                      & 75.88            & -            \\
U$^2$PL \cite{wang2022semi}\ \textcolor[HTML]{9B9B9B}{\scriptsize[{CVPR'22}]} &ResNet-101     & 67.98                   & 69.15            & 73.66           & 76.16                     & 79.49                  \\
GTA-Seg \cite{jin2022semi}\ \textcolor[HTML]{9B9B9B}{\scriptsize[{NeurIPS'22}]} &ResNet-101     & 70.02              & 73.16               & 75.57              & 78.37                  & 80.47                     \\
DAW \cite{sun2024daw}\ \textcolor[HTML]{9B9B9B}{\scriptsize[{NeurIPS'23}]} &ResNet-101     & 74.8            & 77.4                  & 79.5                 & 80.6                   & 81.5                        \\
ESL \cite{ma2023enhanced}\ \textcolor[HTML]{9B9B9B}{\scriptsize[{ICCV'23}]} &ResNet-101    & 70.97                     & 74.06                    & 78.14                 & 79.53                   & 81.77          \\
LogicDiag \cite{liang2023logic}\ \textcolor[HTML]{9B9B9B}{\scriptsize[{ICCV'23}]} &ResNet-101     & 73.25               & 76.66            & 77.93              & 79.39                   & -                 \\
AugSeg \cite{zhao2023augmentation}\ \textcolor[HTML]{9B9B9B}{\scriptsize[{CVPR'23}]} &ResNet-101     & 71.09                 & 75.45              & 78.8                & 80.33         & 81.36                \\
UniMatch \cite{yang2023revisiting}\ \textcolor[HTML]{9B9B9B}{\scriptsize[{CVPR'23}]} &ResNet-101    & 75.2                & 77.2                    & 78.8             & 79.9              & -                       \\
CorrMatch \cite{sun2024corrmatch}\ \textcolor[HTML]{9B9B9B}{\scriptsize[{CVPR'24}]} &ResNet-101     & 76.4                  & 78.5                  & 79.4                    & 80.6                    & 81.8                  \\
DDFP \cite{wang2024towards}\ \textcolor[HTML]{9B9B9B}{\scriptsize[{CVPR'24}]} &ResNet-101     & 74.95                & 78.01             & 79.51              & 81.21             & 81.96                \\
RankMatch \cite{mai2024rankmatch}\ \textcolor[HTML]{9B9B9B}{\scriptsize[{CVPR'24}]} &ResNet-101     & 75.5                  & 77.6                & 79.8                & 80.7                      & 82.2                    \\
AllSpark \cite{wang2024allspark}\ \textcolor[HTML]{9B9B9B}{\scriptsize[{CVPR'24}]} & SegFormer-B5     & 76.07                 & 78.41                 & 79.77                & 80.75              & 82.12              \\ \hline
\textbf{ConformalSAM(AllSpark)(Ours)}      & SegFormer-B5      & {\color[HTML]{00B0F0} 80.69} & {\color[HTML]{00B0F0} 81.29} & {\color[HTML]{00B0F0} 81.33} & {\color[HTML]{00B0F0} 82.69} & {\color[HTML]{00B0F0} 83.44} \\
\textbf{ConformalSAM(Ours)}               & SegFormer-B5      & {\color[HTML]{FE0000} 81.21} & {\color[HTML]{FE0000} 82.22} & {\color[HTML]{FE0000} 81.84} & {\color[HTML]{FE0000} 83.52} & {\color[HTML]{FE0000} 83.85} \\ \hline
\end{tabular}
\vspace{-0.3cm}
\caption{Comparison with state-of-the-art methods on PASCAL VOC validation set with mIoU results. Methods in \textbf{bold} are proposed in this work. All methods are trained on original VOC train set, which consists of 1,464 images. The best and second-best results are colored \textcolor{red}{red} and {\color[HTML]{00B0F0} blue}, respectively}
\vspace{-0.5cm}
\label{table1}
\end{table*}

\section{Experiments}
\subsection{Datasets and Experimental Details}
\textbf{Datasets.}The PASCAL VOC 2012 dataset \cite{everingham2015pascal}, widely used for SSSS, includes 20 object classes and a background class. It consists of 1464 test, 1449 validation, and 1456 training images. The training set is often expanded to 10,582 images using the Segmentation Boundary Dataset (SBD) \cite{hariharan2011semantic}, following \cite{chen2021semi,wang2022semi}. Labeled training ratios typically include 1/16 (662 images), 1/8 (1323 images), 1/4 (2646 images), and 1/2 (5291 images).
ADE20K \cite{zhou2017scene} is a large-scale scene parsing dataset with 150 categories, containing 25,574 training, 2,000 validation, and 3,000 test images. Labeled subsets are created with 1/128 (158 images), 1/64 (316 images), 1/32 (632 images), 1/16 (1,263 images), and 1/8 (2,526 images) of the training set.

\noindent{\textbf{Experimental Details.} For the calibration of SEEM, we set the mis-coverage rate $\alpha$ as 0.05. For the training of the segmentation model, following \cite{wang2024allspark}, we use SGD as the optimizer for all datasets and apply the poly scheduling to adjust the learning rate: $lr = lr_{\text{init}} \cdot \left( 1 - \frac{t}{T} \right)^{0.9}$, where $lr_{\text{init}}$ is the initial learning rate, $t$ is the current optimization step, and $T$ is the max number of optimization steps. On both datasets, we set the initial learning rate as 0.003. During training, the batch size in Stage I is set to 4, while in Stage II, the batch consists of 2 labeled images and 2 unlabeled images. For PASCAL VOC, we train the model for 80 epochs, with 60 epochs allocated for stage I and 20 for stage II. For ADE20K, the training is set to 40 epochs, of which 30 epochs are trained in stage I and 10 epochs are trained in stage II. Additional experimental details, including results using the higher-performing foundational segmentation model GLAMM\cite{rasheed2024glamm} as a replacement for SEEM, are provided in the Appendix.} 

\begin{table*}[t]
\vspace{-0.3cm}
\renewcommand{\arraystretch}{1.2}
\resizebox{1.0\linewidth}{!}{ 
\begin{tabular}{c|cccccc|cccc}
\cline{1-5} \cline{7-11}
\cellcolor[HTML]{DAE8FC}Method                            & \cellcolor[HTML]{DAE8FC}1/16(662)                    & \cellcolor[HTML]{DAE8FC}1/8(1323)                    & \cellcolor[HTML]{DAE8FC}1/4(2646)                    & \cellcolor[HTML]{DAE8FC}1/2(5291)                    &  & \cellcolor[HTML]{DAE8FC}Method & \cellcolor[HTML]{DAE8FC}1/16(662) & \cellcolor[HTML]{DAE8FC}1/8(1323) & \cellcolor[HTML]{DAE8FC}1/4(2646) & \cellcolor[HTML]{DAE8FC}1/2(5291) \\ \cline{1-5} \cline{7-11} 
{\color[HTML]{000000} Lable-only} & {\color[HTML]{000000} 72.01} & {\color[HTML]{000000} 73.20} & {\color[HTML]{000000} 76.62} & {\color[HTML]{000000} 77.61} &  & U$^2$PL\textsuperscript{\textdagger}\cite{wang2022semi}\ \textcolor[HTML]{9B9B9B}{\scriptsize[{CVPR'22}]} & 77.21                             & 79.01                             & 79.30                             & 80.50                             \\ \cline{1-5}
CutMix-Seg \cite{French2019SemisupervisedSS}\ \textcolor[HTML]{9B9B9B}{\scriptsize[{BMVC'20}]}                       & 72.56  & 72.69  & 74.25  & 75.89  &  & PCR\textsuperscript{\textdagger} \cite{xu2022semi}\ \textcolor[HTML]{9B9B9B}{\scriptsize[{NeurIPS'22}]}  & 78.60  & 80.71  & 80.78  & 80.91  \\
CPS \cite{chen2021semi}\ \textcolor[HTML]{9B9B9B}{\scriptsize[{CVPR'21}]} & 72.18  & 75.83  & 77.55  & 78.64  &  & GTA-Seg\textsuperscript{\textdagger} \cite{jin2022semi}\ \textcolor[HTML]{9B9B9B}{\scriptsize[{NeurIPS'22}]}  & 77.82  & 80.47  & 80.57  & 81.01  \\
ESL \cite{ma2023enhanced}\ \textcolor[HTML]{9B9B9B}{\scriptsize[{ICCV'23}]}  & 76.36  & 78.57  & 79.02  & 79.98  &  & LogicDiag\textsuperscript{\textdagger} \cite{liang2023logic}\ \textcolor[HTML]{9B9B9B}{\scriptsize[{ICCV'23}]}  & 79.65  & 80.24  & 80.62  & 81.00  \\
DGCL \cite{wang2023hunting}\ \textcolor[HTML]{9B9B9B}{\scriptsize[{CVPR'23}]}  & 76.61  & 78.37  & 79.31  & 80.96        &  & DAW\textsuperscript{\textdagger} \cite{sun2024daw}\ \textcolor[HTML]{9B9B9B}{\scriptsize[{NeurIPS'23}]}  & 78.82   & 81.19  & 81.03     & 80.62      \\
AugSeg \cite{zhao2023augmentation}\ \textcolor[HTML]{9B9B9B}{\scriptsize[{CVPR'23}]}  & 77.01  & 78.20  & 78.82  & -    &  & SemiCVT\textsuperscript{\textdagger} \cite{huang2023semicvt}\ \textcolor[HTML]{9B9B9B}{\scriptsize[{CVPR'23}]}   & 78.20    & 79.95       & 80.20   & 80.92       \\
UniMatch \cite{yang2023revisiting}\ \textcolor[HTML]{9B9B9B}{\scriptsize[{CVPR'23}]}  & 78.1   & 78.4   & 79.2   & -    &  & CCVC\textsuperscript{\textdagger} \cite{wang2023conflict}\ \textcolor[HTML]{9B9B9B}{\scriptsize[{CVPR'23}]}  & 76.8     & 79.4        & 79.6    & -           \\
CorrMatch \cite{sun2024corrmatch}\ \textcolor[HTML]{9B9B9B}{\scriptsize[{CVPR'24}]}  & 78.4    & 79.3   & 79.6   & -    &  & AugSeg\textsuperscript{\textdagger} \cite{zhao2023augmentation}\ \textcolor[HTML]{9B9B9B}{\scriptsize[{CVPR'23}]}  & 79.29 & 81.46 & 80.50  & -   \\
DDFP \cite{wang2024towards}\ \textcolor[HTML]{9B9B9B}{\scriptsize[{CVPR'24}]}    & 78.32   & 78.88   & 79.83   & 80.90  &  & UniMatch\textsuperscript{\textdagger} \cite{yang2023revisiting}\ \textcolor[HTML]{9B9B9B}{\scriptsize[{CVPR'23}]} & 80.94  & 81.92  & 80.41  & -  \\
RankMatch \cite{mai2024rankmatch}\ \textcolor[HTML]{9B9B9B}{\scriptsize[{CVPR'24}]}  & 78.9  & 79.2  & 80.0  & -   &  & CorrMatch\textsuperscript{\textdagger} \cite{sun2024corrmatch}\ \textcolor[HTML]{9B9B9B}{\scriptsize[{CVPR'24}]}  & 81.3  & 81.9  & 80.9  & -          \\
AllSpark \cite{wang2024allspark}\ \textcolor[HTML]{9B9B9B}{\scriptsize[{CVPR'24}]}  & 78.32  & 79.98  & {\color[HTML]{00B0F0} 80.42}  & {\color[HTML]{00B0F0} 81.14}                                                                      &  & AllSpark\textsuperscript{\textdagger} \cite{wang2024allspark}\ \textcolor[HTML]{9B9B9B}{\scriptsize[{CVPR'24}]}  & 81.56  & 82.04  & 80.92  & {\color[HTML]{00B0F0} 81.13}   \\ \cline{1-5} \cline{7-11} 
\textbf{ConformalSAM(AllSpark)}                                    & {\color[HTML]{FE0000} 80.24}                         & {\color[HTML]{00B0F0} 80.41}                         & 80.04                                                & 79.98                                                &  & \textbf{ConformalSAM(AllSpark)}\textsuperscript{\textdagger}          & {\color[HTML]{00B0F0} 82.34}      & {\color[HTML]{00B0F0} 83.31}      & {\color[HTML]{00B0F0} 81.36}      & 80.57                             \\
\textbf{ConformalSAM}                                              & {\color[HTML]{00B0F0} 80.09}                         & {\color[HTML]{FE0000} 80.93}                         & {\color[HTML]{FE0000} 81.04}                         & {\color[HTML]{FE0000} 81.43}                         &  & \textbf{ConformalSAM}\textsuperscript{\textdagger}                   & {\color[HTML]{FE0000} 83.40}      & {\color[HTML]{FE0000} 83.75}      & {\color[HTML]{FE0000} 82.14}      & {\color[HTML]{FE0000} 81.98}      \\ \cline{1-5} \cline{7-11} 
\end{tabular}
}
\vspace{-0.2cm}
\caption{\textnormal{Comparison with state-of-the-art methods on PASCAL VOC 2012 with augmented training set \cite{hariharan2011semantic}. Methods in \textbf{bold} are proposed in this work. All methods are trained on augmented VOC train set, which consists of 10,582 images. \textsuperscript{\textdagger} means using U\textsuperscript{2}PL’s splits. The best and second-best results are colored \textcolor{red}{red} and {\color[HTML]{00B0F0} blue}, respectively.} \textbf{Left:} Results using the standard splits. \textbf{Right:} Results using U\textsuperscript{2}PL’s splits.}
\vspace{-0.3cm}
\label{table2}
\end{table*}
\noindent{\textbf{Baselines.} We propose two ViT-based baseline models built on SegFormer-B5 \cite{xie2021segformer}. One is a supervised model trained only on labeled data, called \textbf{label-only}. One is the pseudo-label-based approach we mentioned in Section \ref{decay} as our semi-supervised baseline, called \textbf{Semi-Baseline}.}

\begin{table}[]
\renewcommand{\arraystretch}{1.2}
\resizebox{1.0\linewidth}{!}{ 
\begin{tabular}{c|ccccc}
\hline
\rowcolor[HTML]{DAE8FC} 
Method                                     & \begin{tabular}[c]{@{}c@{}}1/128\\ (158)\end{tabular} & \begin{tabular}[c]{@{}c@{}}1/64\\ (316)\end{tabular} & \begin{tabular}[c]{@{}c@{}}1/32\\ (632)\end{tabular} & \begin{tabular}[c]{@{}c@{}}1/16\\ (1263)\end{tabular} & \begin{tabular}[c]{@{}c@{}}1/8\\ (2526)\end{tabular} \\ \hline

Label-only                         & 13.77                                                 & 18.83                                                 & 24.77                                                 & 28.76                                                 & 32.67                                                \\
\hline
CutMix-Seg \cite{French2019SemisupervisedSS} \textcolor[HTML]{9B9B9B}{\scriptsize[{BMVC'20}]} & -                                                     & -                                                     & 26.2                                                  & 29.8                                                  & 35.6                                                 \\
AEL \cite{hu2021semi} \textcolor[HTML]{9B9B9B}{\scriptsize[{NeurIPS'21}]}  & -                                                     & -                                                     & 28.4                                                  & {\color[HTML]{00B0F0} 33.2}                           & {\color[HTML]{FE0000} 38.0}                          \\
UniMatch \cite{yang2023revisiting} \textcolor[HTML]{9B9B9B}{\scriptsize[{CVPR'23}]}  & 15.6                                                  & 21.6                                                  & 28.1                                                  & 31.5                                                  & 34.6                                                 \\
AllSpark \cite{wang2024allspark} \textcolor[HTML]{9B9B9B}{\scriptsize[{CVPR'24}]}  & 16.17                                                 & 23.03                                                 & 26.42                                                 & 28.40                                                 & 32.10                                                \\ \hline
\textbf{ConformalSAM(AllSpark)}      & {\color[HTML]{00B0F0} 24.10}                          & {\color[HTML]{00B0F0} 27.26}                          & {\color[HTML]{00B0F0} 31.40}                          &  32.12                                                & 32.67       \\
\textbf{ConformalSAM}               & {\color[HTML]{FE0000} 26.21}                          & {\color[HTML]{FE0000} 30.02}                          & {\color[HTML]{FE0000} 33.33}                          & {\color[HTML]{FE0000} 34.64}                          & {\color[HTML]{00B0F0} 36.25}                         \\ \hline
\end{tabular}
}
\vspace{-0.3cm}
\caption{\textnormal{Comparison with state-of-the-art methods on ADE20K with mIoU results. Methods in \textbf{bold} are proposed in this work. The best and second-best results are colored \textcolor{red}{red} and {\color[HTML]{00B0F0} blue}, respectively.}}
\vspace{-0.6cm}
\label{table3}
\end{table}

\subsection{Comparison with Existing Methods}
\noindent{\textbf{PASCAL VOC 2012 original.} Tab.~\ref{table1} presents the quantitative results of our method on the PASCAL VOC 2012 original dataset across various label quantities. Compared with state-of-the-art methods, ConformalSAM consistently achieves superior performance across different label amounts. Specifically, compared with AllSpark, our method achieves a 5.14 mIOU improvement on 1/16 (92) labeled data and a 1.73 mIOU increase with Full(1464) labels. Moreover, our method also shows significant improvements when combined with AllSpark. Across the five training settings, ConformalSAM(AllSpark) achieves an average performance gain of 2.46 mIOU compared to the AllSpark. This underscores the efficacy of ConformalSAM and positions it as a flexible framework capable of seamless integration with existing methods.}

\noindent{\textbf{PASCAL VOC 2012 augmented.} The left panel of Tab.~\ref{table2} reports the performance of our method on the augmented training set. Compared to AllSpark, ConformalSAM achieves an average improvement of 0.91 mIOU across all data splits. However, it is notable that ConformalSAM(AllSpark) shows limited performance in the 1/4 and 1/2 partitions, with ConformalSAM yielding only marginal gains. This may be attributed to the presence of partially coarse annotations within the labeled data \cite{wang2024allspark}, which may affect model calibration and reduce pseudo-label quality. Furthermore, under the partitioning scheme of U$^2$PL \cite{wang2022semi} (presented on the right side of Tab.~\ref{table2}), where labeled data includes high-quality annotations, our method achieves more substantial performance improvements, with an average gain of 1.41 mIOU across all partitions. In this setting, ConformalSAM(AllSpark) demonstrates more pronounced benefits.}

\noindent{\textbf{ADE20K.} We evaluated our method on the more challenging ADE20K dataset with results presented in Tab.~\ref{table3}. Compared to previous state-of-the-art models, ConformalSAM achieves mIOU improvements of 10.04, 6.99, 4.93, and 1.44 across the first four labeled partitions, respectively. Notably, in the 1/8 partition, our method demonstrates a significant performance gain over both AllSpark and Label-only. Compared to AllSpark, ConformalSAM(AllSpark) improves 3.1 mIoU on average across all splits, demonstrating the effectiveness of our proposed framework.}

\subsection{Ablation Studies}
\begin{table}[t]
\renewcommand{\arraystretch}{1.2}
\resizebox{\linewidth}{!}{ 
\begin{tabular}{ccc|cccc}
\hline
\multicolumn{3}{c|}{\textit{ConformalSAM}} &                              &                              &                              &                              \\
SEEM          & CP           & SR         & \multirow{-2}{*}{1/16(92)}   & \multirow{-2}{*}{1/8(183)}   & \multirow{-2}{*}{1/4(366)}   & \multirow{-2}{*}{1/2(732)}   \\ \hline
           &         &         & 52.89                        & 69.35                        & 72.37                        & 74.22                        \\
\Checkmark           &         &         & 42.00                        & 41.90                        & 42.37                        & 44.99                        \\
\Checkmark           &          & \Checkmark          & 65.94                        & 70.67                        & 70.52                        & 71.89                        \\
\Checkmark           & \Checkmark          &           & {\color[HTML]{00B0F0} 78.09} & {\color[HTML]{00B0F0} 78.23} & {\color[HTML]{00B0F0} 78.32} & {\color[HTML]{00B0F0} 79.10} \\
\Checkmark           & \Checkmark          & \Checkmark          & {\color[HTML]{FE0000} 81.21} & {\color[HTML]{FE0000} 82.22} & {\color[HTML]{FE0000} 81.84} & {\color[HTML]{FE0000} 83.52} \\ \hline
\end{tabular}

}
\vspace{-0.3cm}
\caption{Ablation study on components of ConformalSAM on the original PASCAL VOC dataset with SegFormer-B5. We use the self-training as our baseline and evaluate the impact of each component, SEEM, CP, and SR under different labeled ratios to evaluate the performance. The best and second-best results are colored \textcolor{red}{red} and {\color[HTML]{00B0F0} blue}, respectively.}
\vspace{-0.3cm}
\label{ablation1}
\end{table}
\noindent\textbf{Effect of the ConformalSAM Module.} Tab.~\ref{ablation1} presents an analysis of the effectiveness of each component within ConformalSAM. We use a simple self-training semi-supervised algorithm as our baseline (the first row). The result shows that using SEEM alone to generate pseudo-labels significantly affects model outputs (the second row). CP leverages SEEM’s knowledge to improve the quality of pseudo-labels generated by SEEM, yielding improvements of 25.2, 8.88, 5.95 and 4.88 mIOU across all label proportions, consistently surpassing the baseline. Additionally, in order to ameliorate the problem of a small amount of noise in the calibrated pseudo-labels, we drop the SEEM-generated masks in Stage II. Specifically, self-reliance (SR) training brings 3.12, 3.99, 3.52 and 4.42 mIoU improvement to the model under different splits, respectively. Furthermore, we also provide the visual validation of pseudo-labels for some samples in Fig.~\ref{fig:visualization}. The pseudo-labels generated by CP are more complete and accurate targets compared to the previous ones, but there is a small amount of noise in the figure. The model being trained provides more accurate pseudo-labels and sufficiently exploits training data of the target domain.

\begin{figure}[]
    \centering
    \begin{subfigure}{0.236\textwidth}
        \includegraphics[width=\linewidth]{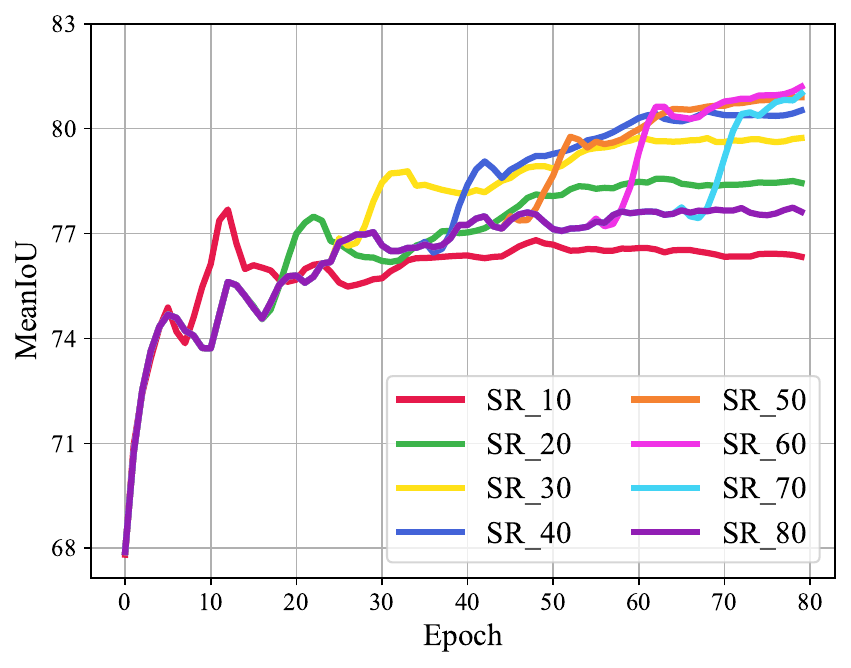}
        \caption*{(a). 1/16 (92)}
    \end{subfigure}
    \centering
    \begin{subfigure}{0.236\textwidth}
        \includegraphics[width=\linewidth]{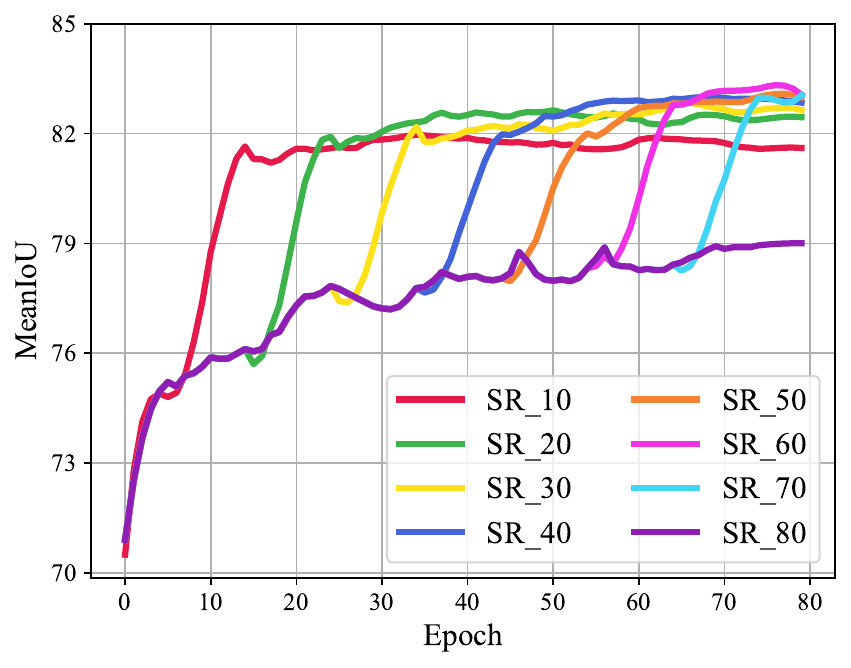}
        \caption*{(b). 1/2 (732)}
    \end{subfigure}
    \vspace{-0.6cm}
    \caption{Ablation study of SR training starting from different epochs on PASCAL VOC original dataset with 1/16(92) and 1/2(732) as the labeled data ratio.}
    \vspace{-0.3cm}
    \label{fig:92_732.pdf}
\end{figure}

\begin{table}[]
\centering
\small
\begin{tabular}{c|c|cc}
\hline
Conformal Calibrated     & Error Rate & \multicolumn{2}{c}{PASCAL VOC 2012} \\
Variants                 & $\alpha$   & 1/16(92)         & 1/2(732)         \\ \hline
\multirow{3}{*}{Pixel}   & 0.1        & 74.31             & 75.95            \\
                         & 0.05       & {\color[HTML]{FE0000} 78.09}            & {\color[HTML]{FE0000} 79.10}            \\
                         & 0.01       & 68.01             & 54.04             \\ \hline
\multirow{3}{*}{Image}   & 0.1        & 75.99             & 77.47             \\
                         & 0.05       & 75.54             & 76.27             \\
                         & 0.01       & 44.59             & 44.10             \\ \hline
\multirow{3}{*}{K-Means} & 0.1        & 69.36             & 78.23             \\
                         & 0.05       & 69.13             & 77.02             \\
                         & 0.01       & 44.16             & 38.71             \\ \hline
\multirow{3}{*}{GenAnn} & 0.1        & 75.41             & 77.25        \\
                         & 0.05       & 73.72             & 75.88            \\
                         & 0.01       & 46.80             & 43.44             \\ \hline
\end{tabular}
\vspace{-0.3cm}
\caption{Ablation study on different conformal calibration variants and error rates. We conducted experiments on the original PASCAL VOC dataset using four different calibration methods (Pixel, Image, K-Means and GenAnn) and three different error rates (0.1, 0.05, and 0.01). The best results are colored \textcolor{red}{red}.}
\vspace{-0.5cm}
\label{ablation2}
\end{table}

\noindent\textbf{The Necessity of an Appropriate Error Rate in ConformalSAM.} CP is the core of ConformalSAM, and the mis-coverage rate $\alpha$ is a key factor affecting the calibration results. As shown in Tab.~\ref{ablation2}, we conducted experiments on three different error rates based on Pixel's calibration approach, showing that $\alpha$=0.05 is consistently the optimal hyperparameter setting. When $\alpha$ gets larger, e.g. $\alpha=0.1$, it usually leads to fewer pseudo-labels and discontinuous boundaries. In contrast, a smaller $\alpha$ leads to more noisily labeled masks. 

\noindent\textbf{The Effect of Different CP Variants.} We conducted experiments in ConformalSAM using different CP variants in \cite{brunekreef2024kandinsky}, including Image, K-Means and GenAnn. The experimental results of these methods on the original PASCAL VOC dataset for three different error rates are shown in Tab.~\ref{ablation2}. Overall, the pixel-wise CP we used has significant advantages with the better segmentation performance. The pixel-wise calibration approach improves the reliability and adaptability of the image segmentation model through fine-grained uncertainty quantification and provides more high-quality pseudo-masks. On the PASCAL VOC dataset, we evaluate the performance of the vanilla CP method without using class-conditional label selection. When class-conditional label selection is applied, the average performance improves by 34.11 mIoU across the four different labeled data ratios, compared to the vanilla CP method.

\noindent\textbf{The Optimal Time of Starting SR Training.} Tab.~\ref{ablation1} shows the SR strategy in Stage II improves the final performance substantially. However, choosing the right moment for the transition from Stage I to Stage II is equally important. Fig.~\ref{fig:92_732.pdf} shows our experiments with different epochs to start SR training when there are 92 and 732 labeled images. Both early SR (10 epochs) and late SR (70 epochs) significantly improve performance, but early SR of the model may result in insufficient learning from SEEM and overfitting in the later training stage, ultimately diminishing model performance. This overfitting issue is particularly pronounced when the labeled data is limited. Overall, our findings indicate that transitioning to SR at 60 epochs yields the best results.

\noindent\textbf{Limitations.} As stated in the introduction, our work aims to unleash the strong capability of a foundation model on downstream tasks by calibrating the model output with CP. Therefore, the effectiveness of ConformalSAM relies on the degree of overlap between the foundation model’s knowledge and the target downstream task. For datasets with substantial overlap, such as PASCAL VOC 2012, CP enables reliable confidence control. However, in datasets like ADE20K and Cityscapes \cite{cordts2016cityscapes} (we report experimental results on Cityscapes in the Supplementary Material, where ConformalSAM achieves performance comparable to that of AllSpark), which contain many novel categories, the foundation model’s prior knowledge is less applicable to these unseen classes. Despite this challenge, ConformalSAM maintains competitive performance compared to state-of-the-art methods. As foundation models continue to expand their knowledge, we believe this limitation will gradually become less relevant, allowing ConformalSAM to become even more valuable across diverse downstream tasks due to its flexibility. 
\section{Conclusion}
\label{sec:conclusion}
This work investigates the research question: can predictions from a foundational segmentation model be directly used as supervision signals for unlabeled data in SSSS? Our empirical study reveals that directly using SEEM-generated masks even worsens the performance compared with only using the labeled images. To obtain reliable masks, we design ConformalSAM to filter out unreliable pixel labels using CP as the uncertainty probe with labeled images. To avoid overly relying on the SEEM-generated masks, we further design the self-reliance training strategy that drops out the SEEM labels at the later stage of training. On three standard SSSS datasets, ConformalSAM achieves the best performance when compared with strong baselines and also improves the performance of a recent strong SSSS method as a plug-in. Our ConformalSAM reveals the potential of using foundational segmentation models as annotators when used properly and our future work will explore the effectiveness of uncertainty calibration in other foundation models such as large language models.

{
    \small
    \bibliographystyle{ieeenat_fullname}
    \bibliography{ref}
}


\end{document}